\documentclass[runningheads]{llncs}
\usepackage[T1]{fontenc}
\usepackage{graphicx}
\usepackage{booktabs}
\usepackage[misc]{ifsym}
\usepackage{multirow}
\usepackage{tabularx}
\usepackage{amsfonts}
\usepackage{subcaption}
\usepackage[export]{adjustbox}
\usepackage{bbm}
\newcommand{\corr}{(\Letter)}
\usepackage{mwe}

\begin{document}

\title{Applications of temporal graph learning for predicting the dynamics of biological systems}

\titlerunning{Temporal graph learning for biology}

\author{Manuel Dileo\inst{1}\orcidID{0000-0002-4861-455X} \corr \and
Andrea Sottoriva\inst{1} \orcidID{0000-0001-6709-9533}}

\authorrunning{M. Dileo et al.}

\institute{Center for Computational Biology, Human Technopole Foundation, Milan, Italy \email{\{manuel.dileo,andrea.sottoriva\}@fht.org}}

\maketitle              

\begin{abstract}
Recent biological foundation models have shown strong performance in single-cell representation learning by applying transformer architectures directly to gene-expression matrices. However, these approaches predominantly operate in static settings and do not explicitly model the temporal evolution of developmental programs in the cell. Modeling such dynamics is important for understanding how cellular states progressively emerge, differentiate, and reorganize during development or disease progression. In this work-in-progress paper, we investigate an alternative temporal graph-based perspective in which cellular states are represented through pseudotime-resolved gene regulatory networks and modeled as evolving graph structures over persistent gene identities. Starting from single-cell transcriptomic data, we infer pseudotime trajectories, discretize cells into developmental snapshots, reconstruct one gene regulatory network per snapshot, and apply temporal graph neural networks to forecast future biological states. We evaluate this framework on two publicly available mouse developmental datasets, erythroid gastrulation and pancreatic endocrinogenesis, considering three complementary tasks: gene-expression forecasting, future link prediction, and future out-degree centrality prediction. Our results show that graph-based models outperform well-known biological foundation-model such as scGPT and scFoundation, suggesting that explicitly modeling evolving regulatory structure provides useful information beyond static pretrained representations. For link prediction and centrality forecasting, temporal graph learning captures non-trivial regulatory dynamics and enables the identification of temporally important gene hubs. Overall, our findings support temporal graph learning as a promising direction for modeling dynamic biological systems and as a complementary paradigm to current foundation model approaches in single-cell biology.

\end{abstract}

\section{Introduction}

Recent advances in single-cell transcriptomics and biological foundation models have significantly improved the ability to learn representations of genes, cells, and cellular states from large-scale transcriptomic datasets. In particular, transformer-based architectures such as scGPT~\cite{scgpt}, scFoundation~\cite{scfoundation}, and Geneformer~\cite{geneformer} have shown strong performance across a broad range of downstream tasks, including cell annotation, perturbation prediction, and gene-expression imputation. These approaches typically operate directly on gene-expression matrices and learn contextualized representations through masked reconstruction objectives analogous to masked language modeling. Despite these advances, most existing approaches remain fundamentally static and do not explicitly model the temporal evolution of cells. This limitation is particularly relevant in developmental biology, where cellular states continuously emerge, differentiate, and reorganize over time and during diseas progression. Modeling such dynamics remains challenging because standard single-cell RNA sequencing technologies are inherently destructive: each cell can only be observed once, preventing direct temporal tracking of individual cellular trajectories. As a result, developmental progression must instead be reconstructed computationally from large populations of asynchronously observed cells. Recent advances in pseudotime inference and RNA velocity analysis have partially addressed this limitation by enabling the reconstruction of ordered developmental trajectories from single-cell transcriptomic data~\cite{scvelo,dpt}. However, most existing methods still focus on modeling cellular states independently and do not explicitly represent the evolving regulatory structure underlying developmental transitions.
At the same time, recent progress in temporal graph learning has shown that temporal graph neural networks (TGNNs) can effectively model evolving relational systems by jointly capturing structural dependencies and temporal dynamics across graph snapshots~\cite{longa23tgl}. This naturally raises the question of whether developmental biological systems can be reformulated as evolving graph processes suitable for temporal graph learning. 

In this work-in-progress paper, we investigate temporal graph learning as a framework for modeling and predicting the evolution of biological systems from single-cell transcriptomic data. Starting from single-cell gene-expression matrices, we first reconstruct developmental trajectories through diffusion pseudotime \cite{dpt}, discretize cells into developmental snapshots, and infer one gene regulatory network (GRN) per pseudotime bin using SCENIC~\cite{scenic}. This procedure transforms developmental single-cell systems into discrete-time temporal graphs with persistent gene identities and evolving regulatory interactions. Temporal graph neural networks are then applied to learn temporal node representations and forecast future biological states. We evaluate this framework on two publicly available mouse developmental datasets, erythroid gastrulation and pancreatic endocrinogenesis, considering three complementary forecasting tasks: future gene-expression prediction, future link prediction, and future out-degree centrality prediction. Our results show that graph-based temporal models outperform widely used biological foundation models such as scGPT and scFoundation on gene expression forecasting, suggesting that explicitly modeling evolving regulatory structure provides useful information beyond static pretrained representations. Interestingly, spectral graph architectures such as ChebNet and GCRN-GRU consistently emerge as some of the strongest-performing models across tasks, suggesting that spectral graph convolutions may be particularly well suited for the heterophilous and hub-centered structure of gene regulatory networks. 

Overall, our findings support temporal graph learning as a promising and complementary paradigm to current foundation-model approaches for modeling dynamic biological systems.

\section{Background}
\label{sec:background}

\paragraph{Virtual cell modeling.}
Recent advances in foundation models for single-cell biology have stimulated growing interest in \emph{virtual cell} modeling, namely the development of computational systems capable of simulating cellular states, perturbations, and transcriptional responses~\cite{virtualcell,state}. 
Most existing approaches are based on large pretrained transformer architectures trained through masked gene reconstruction objectives or perturbation prediction tasks~\cite{scgpt,scfoundation}. 
These models primarily focus on learning contextualized representations of genes and cells from large-scale transcriptomic corpora, enabling downstream applications such as cell-state annotation, perturbation prediction, and gene-expression imputation. Graph-based approaches such as GEARS~\cite{gears} and HEIST~\cite{madhu2026heist} have introduced explicit modeling of gene interactions. However, these methods still operate in essentially static settings and do not explicitly model the temporal evolution of regulatory programs across developmental trajectories. In contrast, temporal graph learning provides a natural framework for modeling evolving biological systems.

\paragraph{Temporal Graph Neural Networks.}
Temporal graph neural networks (TGNNs) are deep learning models designed to learn, predict, and reason over evolving relational structures. A comprehensive overview of the main architectures and design choices is provided in recent surveys~\cite{longa23tgl}. Existing TGNN approaches are commonly categorized according to how temporal information is represented. In continuous-time models, dynamic graphs are treated as streams of timestamped events, whereas in discrete-time or snapshot-based approaches, temporal networks are represented as ordered sequences of static graphs observed at a chosen temporal granularity~\cite{tgb}. These two formulations target different types of temporal dynamics and lead to substantially different modeling assumptions and evaluation protocols. Event-based approaches are particularly suitable for high-frequency interaction streams and fine-grained temporal reasoning tasks, whereas snapshot-based approaches are better aligned with systems evolving through relatively coherent temporal phases~\cite{evolvegcn}. In this work, we focus on discrete-time TGNNs, since developmental biological systems naturally evolve through progressive transcriptional states that can be represented as ordered pseudotime-resolved snapshots. Among discrete-time TGNNs, some of the most effective architectures combine graph message passing with explicit temporal update mechanisms. Common design strategies include parameter evolution~\cite{evolvegcn}, recurrent embedding evolution~\cite{gconvgru}, and temporal memory modules inspired by sequence modeling\cite{roland}. Alternative lines of work incorporate temporal point process (TPP) components to model event intensities and temporal dependencies, offering a complementary perspective on temporal dynamics that can be adapted to discretized settings \cite{Qi2025Fusion}. Overall, these approaches highlight the importance of jointly modeling structural dependencies within snapshots and temporal evolution across snapshots.

\paragraph{Gene Regulatory Networks.}
Gene regulatory networks (GRNs) represent regulatory interactions between genes, transcription factors, and downstream targets, and constitute a fundamental abstraction for studying cellular regulation and developmental processes. In GRNs, nodes correspond to genes, while directed edges encode putative regulatory relationships inferred from transcriptional activity patterns or prior biological knowledge. GRNs have been widely used to study cell differentiation, developmental trajectories, disease progression, and transcriptional control programs \cite{Karlebach2008}. A broad range of computational approaches has been proposed for GRN inference, including correlation-based methods, information-theoretic approaches, regression models, probabilistic graphical models, tree-based ensemble techniques, and deep learning approacches~\cite{SanguinettiGRNInference,DongGRNDeep,DaSilvaGRNSurvey}. More recently, single-cell transcriptomics has enabled the reconstruction of context-specific GRNs at cellular resolution, allowing the study of regulatory programs across heterogeneous cell populations and developmental trajectories. Among the most widely adopted frameworks, SCENIC~\cite{scenic} combines co-expression analysis with transcription factor motif enrichment to infer biologically meaningful regulatory interactions from single-cell RNA sequencing data. Despite the increasing availability of single-cell datasets, the vast majority of existing GRN studies still treat regulatory networks as static objects. In reality, regulatory programs evolve continuously during development, differentiation, and cellular adaptation. Recent advances in pseudotime inference and RNA velocity analysis~\cite{scvelo} now make it possible to reconstruct ordered developmental trajectories from single-cell data, enabling the construction of pseudotime-resolved GRNs that evolve over developmental time. In this work, we leverage this setting to study temporal graph learning over evolving regulatory networks and evaluate whether TGNNs can model the future evolution of biological systems.

\section{Methodology}
\label{sec:method}

\begin{figure}
    \centering
    \includegraphics[width=\linewidth]{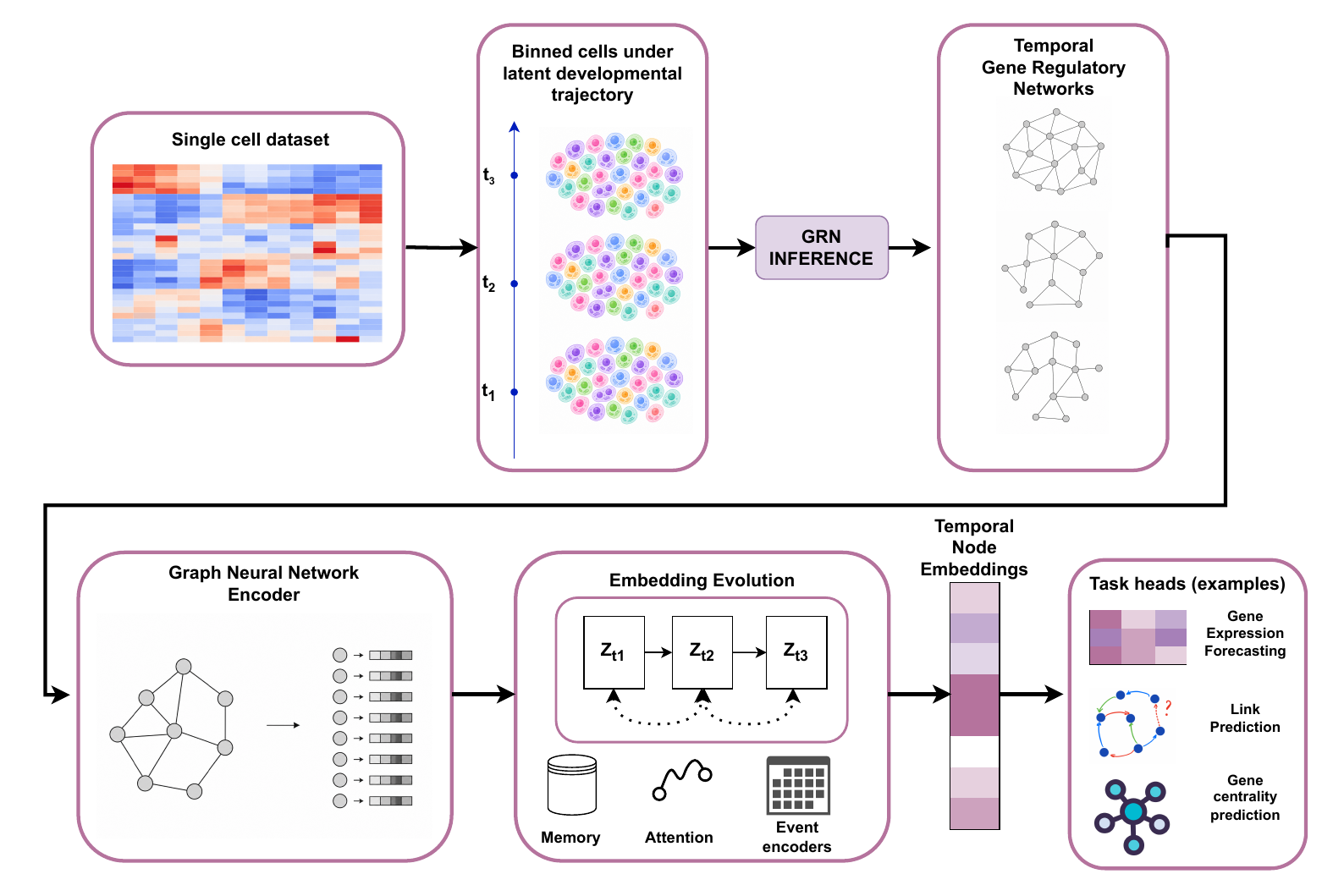}
    \caption{Overview of the proposed temporal graph learning pipeline for modeling and predicting the evolution of biological systems.}
    \label{fig:pipeline}
\end{figure}

Figure~\ref{fig:pipeline} provides an overview of the proposed temporal graph learning pipeline for modeling and predicting the evolution of biological systems.
Starting from single-cell gene-expression matrices, we first infer developmental trajectories through pseudotime estimation \cite{dpt}, obtaining an ordered representation of cellular progression across differentiation states. Cells are subsequently discretized into pseudotime bins, and one gene regulatory network (GRN) is inferred for each temporal bin. This procedure yields a discrete-time temporal GRN represented as an ordered sequence of pseudotime-resolved graph snapshots. Temporal graph neural networks are then applied to the resulting temporal GRNs in order to learn temporal node representations that jointly capture regulatory structure and developmental dynamics across snapshots. The learned temporal embeddings are subsequently used for downstream forecasting tasks, including future link prediction, future gene-expression prediction, and future out-degree centrality prediction.

\paragraph{Temporal GRN construction.}
The input of the proposed pipeline is a single-cell RNA sequencing dataset represented as a gene-expression matrix, where rows correspond to genes, columns correspond to cells, and values represent gene-expression levels. Starting from this representation, we first infer developmental trajectories through diffusion pseudotime (DPT)~\cite{dpt}, which assigns a continuous scalar value to each cell representing its relative developmental progression along the inferred trajectory. Diffusion pseudotime first constructs a neighborhood graph of transcriptionally similar cells and then estimates developmental progression through diffusion distances over this graph. We adopt DPT because it provides biologically coherent temporal orderings of cells across the studied developmental processes (see Appendix). Because pseudotime is defined at the single-cell level whereas GRN inference requires sufficiently large cell populations to estimate gene--gene regulatory dependencies, cells are aggregated within temporal bins before network reconstruction.
Each bin therefore represents a locally coherent developmental state from which one GRN snapshot can be inferred. The number of bins is selected through a trade-off between temporal resolution and statistical robustness of GRN inference. Smaller bins provide finer temporal resolution but may lead to unstable regulatory inference due to insufficient numbers of cells, whereas larger bins improve statistical reliability at the cost of smoothing temporal dynamics. In practice, we typically target approximately at least $300$--$500$ cells per temporal snapshot in order to maintain sufficiently stable GRN inference while avoiding excessively coarse temporal aggregation. For each pseudotime bin, we infer one gene regulatory network (GRN) using SCENIC~\cite{scenic}, one of the most widely adopted frameworks for single-cell GRN inference. SCENIC first identifies candidate regulatory interactions through gene co-expression analysis and subsequently refines them through transcription factor motif enrichment analysis to obtain biologically meaningful regulatory programs. The resulting biological system is modeled as a discrete-time temporal graph
$\mathcal{G} = \{ G^{(1)}, G^{(2)}, \dots, G^{(T)} \}$,
where each snapshot
$G^{(t)} = (V, E^{(t)}, \mathbf{X}^{(t)}, \mathbf{E}^{(t)})$
represents the GRN inferred for pseudotime bin $t$.
Here, $V$ denotes the set of genes,
$E^{(t)} \subseteq V \times V$ the set of regulatory interactions at time $t$,
$\mathbf{X}^{(t)} \in \mathbb{R}^{|V| \times d_x}$ the node feature matrix, and
$\mathbf{E}^{(t)} \in \mathbb{R}^{|E^{(t)}| \times d_e}$ the edge feature matrix. Node features are obtained through aggregation of gene-expression statistics within each pseudotime bin, including mean, median, standard deviation, fraction of non-zero expression values, and total expression. Edge features correspond to the confidence scores produced by the GRN inference procedure. The topology, node features, and edge attributes are therefore all allowed to evolve across developmental pseudotime.

\paragraph{Problem statement.}

Given a temporal gene regulatory network
$\mathcal{G} = \{ G^{(1)}, \dots, G^{(T)} \}$,
our objective is to predict future structural and functional states of the evolving biological system from historical developmental snapshots.

\begin{itemize}
    \item Future gene-expression forecasting: The first task is formulated as temporal node-level regression. Given the historical sequence $\{ G^{(1)}, \dots, G^{(t)} \}$, the objective is to predict future transcriptional changes between consecutive developmental snapshots. Rather than directly forecasting the full node-feature vectors, we specifically predict the temporal variation in mean gene expression between snapshots $t$ and $t+1$. Formally, let $\mathbf{x}^{(t)} \in \mathbb{R}^{|V|}$ denote the vector containing the mean expression value of each gene at time $t$. The task consists of estimating \[ \Delta \hat{\mathbf{x}}^{(t+1)} = \mathbf{x}^{(t+1)} - \mathbf{x}^{(t)}, \] using the temporal history $\mathcal{G}_{\leq t}$. Predicting gene-expression is one of the core objectives underlying modern biological foundation models, which are commonly trained through masked gene-expression reconstruction objectives analogous to masked language modeling~\cite{scgpt,scfoundation,state}. Hence, gene-expression forecasting provides a natural setting to evaluate whether temporal graph learning offers advantages over biological foundation models for modeling dynamic biological systems.
    \item Future link prediction: Given the historical sequence $\{ G^{(1)}, \dots, G^{(t)} \}$, the goal is to predict the presence of regulatory interactions in the subsequent snapshot $G^{(t+1)}$. Formally, for any pair of genes $(i,j) \in V \times V$, the task consists of estimating the probability \[ \hat{y}_{ij}^{(t+1)} = P\big((i,j) \in E^{(t+1)} \mid \mathcal{G}_{\leq t}\big). \] Future link prediction serves as a standard benchmark task in temporal graph learning and provides a useful validation of whether the constructed pseudotime-resolved GRNs contain forecastable temporal structure.
    \item Future node centrality prediction: The third task is formulated as temporal node-level regression. Given the historical sequence $\{ G^{(1)}, \dots, G^{(t)} \}$, the objective is to predict future node centrality values \cite{barabasi2016network} in snapshot $G^{(t+1)}$. Predicting future topological properties of evolving biological networks provides an additional perspective on the capabilities of temporal graph learning beyond edge forecasting and gene-expression prediction. This task illustrates a form of topological forecasting naturally enabled by temporal graph learning but not directly supported by standard tabular models or sequence-based biological foundation models. While transformer-based foundation models primarily operate on gene-expression reconstruction objectives, temporal graph learning allows explicit modeling of the future structural role of genes within evolving regulatory networks. In this work, we focus on future out-degree centrality as one intuitive example of regulatory importance in gene regulatory networks, since genes with high out-degree can be interpreted as putative regulatory hubs controlling multiple downstream targets. Formally, for each gene $i \in V$, the out-degree centrality at time $t+1$ is defined as \[ c_i^{\mathrm{out},(t+1)} = \frac{1}{|V|-1} \sum_{j \in V} \mathbbm{1}\big((i,j) \in E^{(t+1)}\big), \] where $\mathbbm{1}(\cdot)$ denotes the indicator function.
\end{itemize}

\section{Experiments}

\paragraph{Experimental setup.}We evaluate temporal graph learning models on three forecasting tasks over pseudotime-resolved gene regulatory networks: future link prediction, gene-expression forecasting, and future out-degree centrality prediction. All experiments are conducted in the live-update setting~\cite{roland}, which explicitly accounts for the evolving nature of developmental biological systems. Under this protocol, models are first evaluated on the next temporal snapshot and subsequently refined on the newly observed data before proceeding to future timestamps. For future link prediction, we evaluate models using the area under the precision-recall curve (AUPRC), following prior work on temporal graph learning~\cite{Yang2015,RossiTGCN,evolvegcn,poursafaei2022towards}. For each temporal snapshot, negative edges are sampled by pairing genes that are active within the current snapshot but are not connected by regulatory interactions. This strategy provides a more challenging and biologically realistic setting compared to uniformly sampling random node pairs, which may generate trivially disconnected or inactive genes. For future gene-expression forecasting, models predict the future variation in gene expression between consecutive pseudotime snapshots.
We evaluate performance using Pearson correlation coefficient (PCC) \cite{scfoundation} together with Precision@200 for the most upregulated and downregulated genes, allowing us to measure both global predictive accuracy and the ability to recover the strongest transcriptional transitions. Finally, we evaluate future out-degree centrality prediction using mean absolute error (MAE), Spearman correlation, and Precision@200 over the highest-centrality genes. Code, data, and supplementary information about the experiments can be found in our GitHub repository\footnote{\url{https://anonymous.4open.science/r/tgl-grn-1CCD}}.

\paragraph{Models.} We considered a broad set of graph-based, temporal, tabular, and biological foundation-model baselines. As primary temporal graph learning baselines, we selected three widely adopted discrete-time temporal graph neural networks: EvolveGCN~\cite{evolvegcn}, GCRN-GRU~\cite{gconvgru}, and ROLAND~\cite{roland}. These models represent two major paradigms in temporal graph learning, namely parameter evolution and embedding evolution~\cite{longa23tgl}.  To evaluate the contribution of temporal modeling, we additionally considered static graph neural network architectures, namely GCN~\cite{gcn}, GAT~\cite{gat}, and ChebNet~\cite{chebnet}. We also included simple supervised tabular baselines, such as a linear model and a three-layer MLP. To compare against recent large-scale pretrained models for biology, we additionally evaluated scGPT~\cite{scgpt} and scFoundation~\cite{scfoundation}. These models provide pretrained representations of genes and cells learned from large-scale single-cell datasets through self-supervised tasks. A detailed description of all considered models is provided in the Appendix. We plan to extend this benchmark with additional foundation models in biology and more recent temporal attention-based architectures such as DyGFormer~\cite{dygformer}, which were originally proposed for event-based temporal graph modeling and may provide additional advantages in modeling long-range developmental dependencies.

\begin{figure}[t]
\centering

\begin{subfigure}[t]{0.48\textwidth}
\centering
\includegraphics[width=0.95\linewidth]{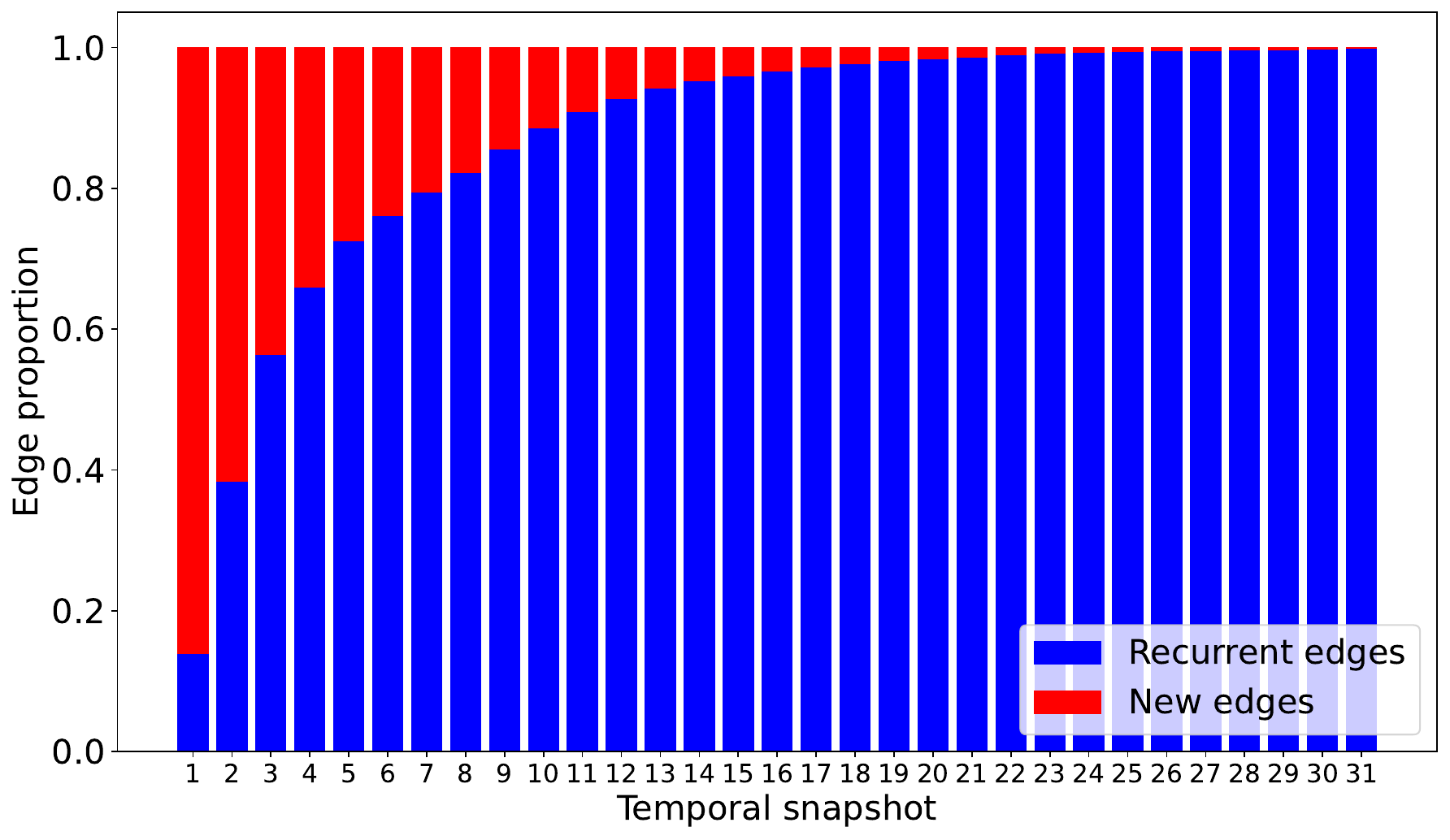}
\caption{}
\end{subfigure}
\hfill
\begin{subfigure}[t]{0.48\textwidth}
\centering
\includegraphics[width=0.95\linewidth]{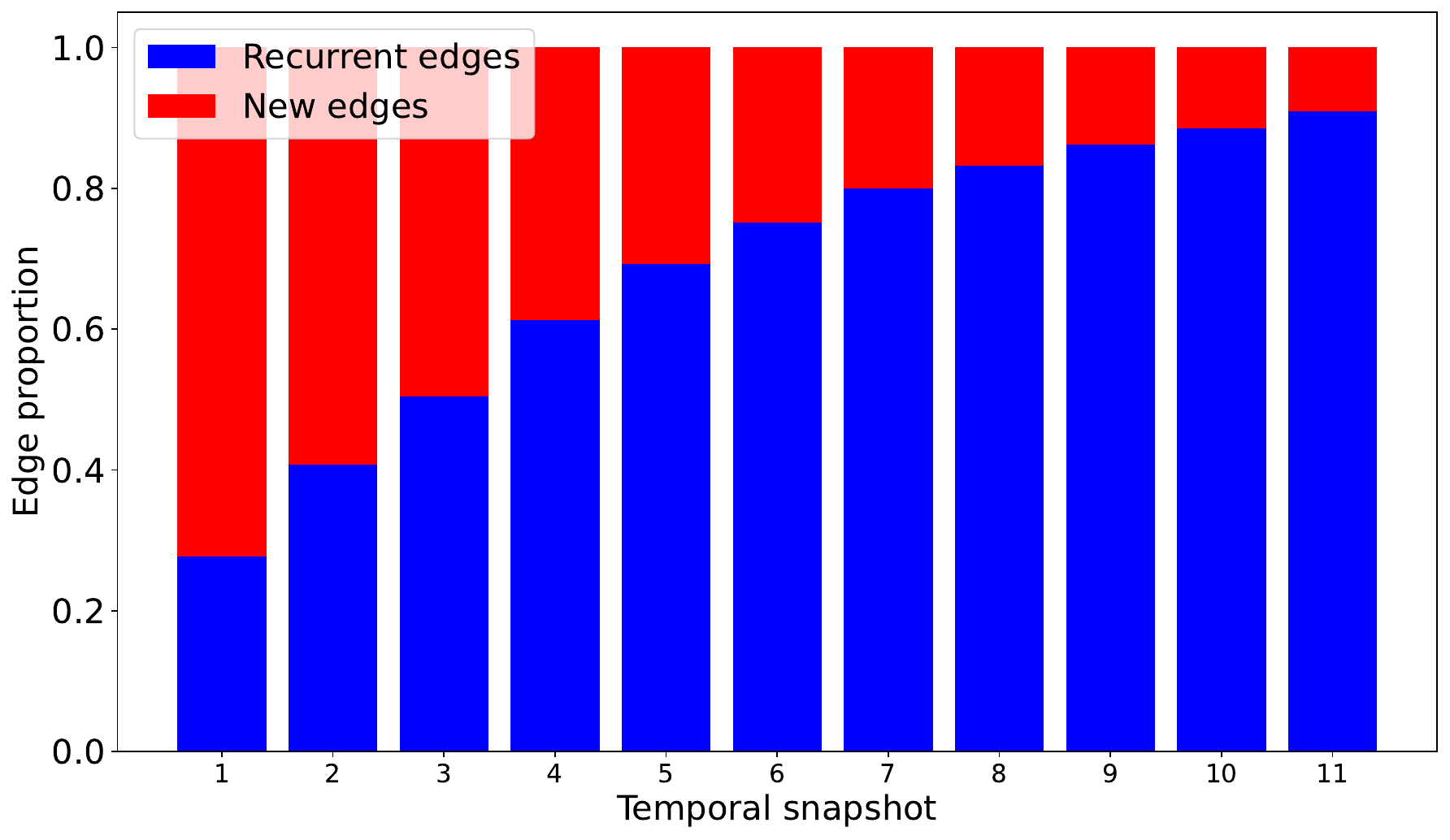}
\caption{}
\end{subfigure}
\caption{Proportion of recurrent vs new edges over time for developmental GRN snapshots of mouse gastrulation (a) and pancreas (b).}
\label{fig:num-edges}
\end{figure}

\paragraph{Datasets.} We evaluated temporal graph learning models on two publicly available single-cell developmental datasets commonly used in the RNA velocity literature ~\cite{scvelo}: mouse erythroid gastrulation~\cite{Pijuan-Sala2019} and mouse pancreas endocrinogenesis~\cite{pancreas}. 
The erythroid gastrulation dataset captures early hematopoietic development during mouse embryogenesis, whereas the pancreas dataset describes endocrine lineage differentiation during pancreatic development. For each dataset, we constructed a temporal GRN following the methodology presented in Section \ref{sec:method}. The resulting datasets contain 32 temporal snapshots for gastrulation and 12 snapshots for pancreas, reflecting the different pseudotime trajectory structures observed in the two developmental systems \cite{scvelo}. We report the main dataset statistics in Table~\ref{tab:dataset}. Overall, both datasets exhibit highly recurrent graph structures over developmental time, with a progressively increasing fraction of recurring regulatory interactions across pseudotime snapshots. The temporal evolution of recurrent and novel edges is visualized in Figure~\ref{fig:num-edges}. In particular, the gastrulation dataset shows that only a relatively small fraction of interactions remain novel after the first half of the developmental trajectory, suggesting that the global regulatory backbone stabilizes rapidly during erythroid commitment while subsequent developmental progression is primarily characterized by gradual rewiring of an already established regulatory structure. Importantly, despite this strong edge recurrence, pure memorization strategies such as EdgeBank still underperform temporal graph learning approaches for future link prediction (see below), indicating that the datasets contain non-trivial temporal dynamics beyond simple edge persistence. Additional dataset analysis are available in the Appendix.

\begin{table}
\caption{Key dataset statistics. Avg recurrence is the averaged proportion of recurrent edges over the snapshots.}
\label{tab:dataset}
\centering
\begin{tabular}{ccccc}
\hline
Dataset & \#Nodes & \#Edges & \#Snapshots & Avg recurrence \\
\hline
mouse-gastrulation & 6,345 & 3,514,167 & 32 & 0.87 \\
mouse-pancreas & 7,197 & 4,153,514 & 12 & 0.68  \\
\hline
\end{tabular}
\end{table}

\paragraph{Forecasting gene expression.} We report the results in Table~\ref{tab:gene_forecasting}. Overall, scGPT and scFoundation generally achieved lower performance than graph-based approaches despite leveraging large-scale biological pretraining, suggesting that static foundation-model representations alone may be insufficient to capture the temporal evolution of regulatory programs. In contrast, temporal graph models obtained the highest average performance, with GCRN-GRU achieving the best Pearson correlation coefficient (PCC) on mouse gastrulation and competitive performance on mouse pancreas. Interestingly, standard supervised tabular baselines such as MLPs achieved competitive PCC values on mouse pancreas, indicating that local expression statistics already contain substantial predictive information.  A notable observation is that the strongest-performing graph architectures were predominantly based on spectral graph convolutions, namely ChebNet and GCRN-GRU. One possible explanation is related to the heterophilous and hub-centered structure of gene regulatory networks, where transcription factors act as central regulators connected to multiple downstream target genes in star-like topologies. Under this structure, spectral convolutions may more effectively propagate information across multi-hop neighborhoods, allowing regulated genes sharing the same transcription factor to exchange information through short graph paths while potentially mitigating some of the limitations commonly associated with standard message-passing architectures \cite{arnaiz-rodriguez2026oversmoothing}. This hypothesis will be investigated more systematically in planned future work  on this project. Finally, the relatively small performance gap between ChebNet and the temporal GCRN-GRU architecture suggests that, in many cases, considering the immediately preceding GRN snapshot may already be sufficient to predict the next developmental state. This observation may indicate that pseudotime-resolved GRN evolution is locally smooth and that long-range temporal memory mechanisms provide only limited additional benefit for these forecasting tasks.

\begin{table}[t]
\centering
\small
\caption{Comparison of methods for gene-expression forecasting on mouse gastrulation and mouse pancreas. For each model, we ran experiments with 5 different random seeds, reporting the average result and standard deviation for each method. Higher PCC and P@200 indicate better performance. Best results are in \textbf{bold}, second best are \underline{underlined}.}
\label{tab:gene_forecasting}
\setlength{\tabcolsep}{4pt}

\resizebox{\textwidth}{!}{
\begin{tabular}{lccc|ccc}
\toprule
& \multicolumn{3}{c|}{mouse-gastrulation}
& \multicolumn{3}{c}{mouse-pancreas} \\
\cmidrule(lr){2-4}
\cmidrule(lr){5-7}

Model
& PCC
& P@200 $\uparrow$
& P@200 $\downarrow$
& PCC
& P@200 $\uparrow$
& P@200 $\downarrow$ \\
\midrule

Linear
& 0.062 $\pm$ 0.04
& 0.041 $\pm$ 0.05
& 0.007 $\pm$ 0.02
& 0.107 $\pm$ 0.05
& 0.058 $\pm$ 0.06
& 0.057 $\pm$ 0.11 \\

MLP
& 0.206 $\pm$ 0.20
& 0.014 $\pm$ 0.03
& 0.063 $\pm$ 0.14
& \textbf{0.434 $\pm$ 0.09}
& 0.098 $\pm$ 0.09
& \underline{0.350 $\pm$ 0.11} \\

\midrule

scFoundation
& 0.100 $\pm$ 0.16
& 0.040 $\pm$ 0.06
& 0.043 $\pm$ 0.08
& 0.308 $\pm$ 0.12
& \textbf{0.178 $\pm$ 0.15}
& 0.127 $\pm$ 0.13 \\

scGPT
& 0.084 $\pm$ 0.09
& 0.014 $\pm$ 0.04
& 0.006 $\pm$ 0.02
& 0.319 $\pm$ 0.13
& 0.084 $\pm$ 0.07
& 0.208 $\pm$ 0.16 \\

\midrule

GCN
& 0.336 $\pm$ 0.12
& 0.080 $\pm$ 0.05
& 0.323 $\pm$ 0.17
& 0.278 $\pm$ 0.13
& 0.083 $\pm$ 0.05
& 0.182 $\pm$ 0.10 \\

GAT
& 0.363 $\pm$ 0.14
& 0.046 $\pm$ 0.05
& 0.335 $\pm$ 0.17
& 0.368 $\pm$ 0.11
& 0.079 $\pm$ 0.05
& 0.319 $\pm$ 0.10 \\

ChebNet
& \underline{0.397 $\pm$ 0.11}
& 0.082 $\pm$ 0.05
& \textbf{0.387 $\pm$ 0.14}
& 0.423 $\pm$ 0.11
& 0.087 $\pm$ 0.05
& \textbf{0.354 $\pm$ 0.12} \\

\midrule

EvolveGCN
& 0.346 $\pm$ 0.12
& 0.084 $\pm$ 0.04
& 0.328 $\pm$ 0.18
& 0.320 $\pm$ 0.11
& 0.099 $\pm$ 0.07
& 0.220 $\pm$ 0.14 \\

GCRN-GRU
& \textbf{0.400 $\pm$ 0.11}
& \underline{0.086 $\pm$ 0.05}
& \underline{0.378 $\pm$ 0.15}
& \underline{0.425 $\pm$ 0.08}
& 0.099 $\pm$ 0.05
& 0.333 $\pm$ 0.13 \\

ROLAND (ChebConv+ GRU)
& 0.134 $\pm$ 0.09
& \textbf{0.110 $\pm$ 0.09}
& 0.084 $\pm$ 0.07
& 0.219 $\pm$ 0.08
& \underline{0.173 $\pm$ 0.10}
& 0.124 $\pm$ 0.09 \\

\bottomrule
\end{tabular}
}
\end{table}

\begin{figure}[t]
\centering

\begin{subfigure}[t]{0.48\textwidth}
\centering
\includegraphics[width=0.95\linewidth]{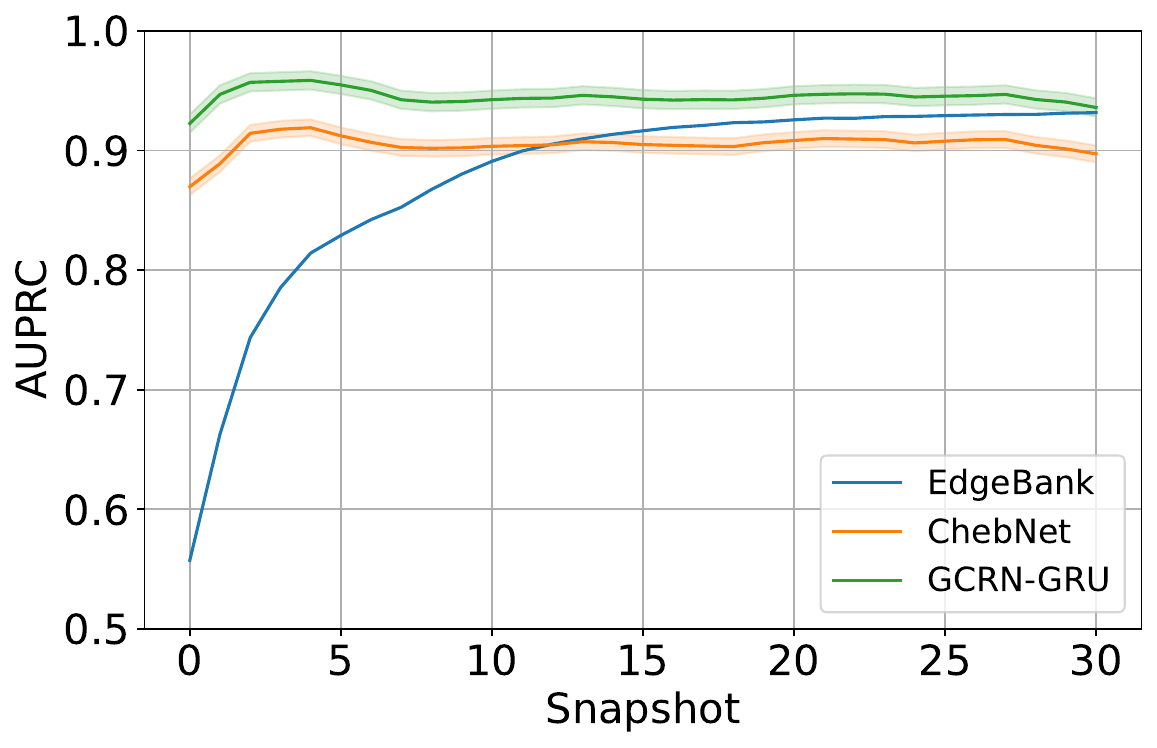}
\caption{}
\end{subfigure}
\hfill
\begin{subfigure}[t]{0.48\textwidth}
\centering
\includegraphics[width=0.95\linewidth]{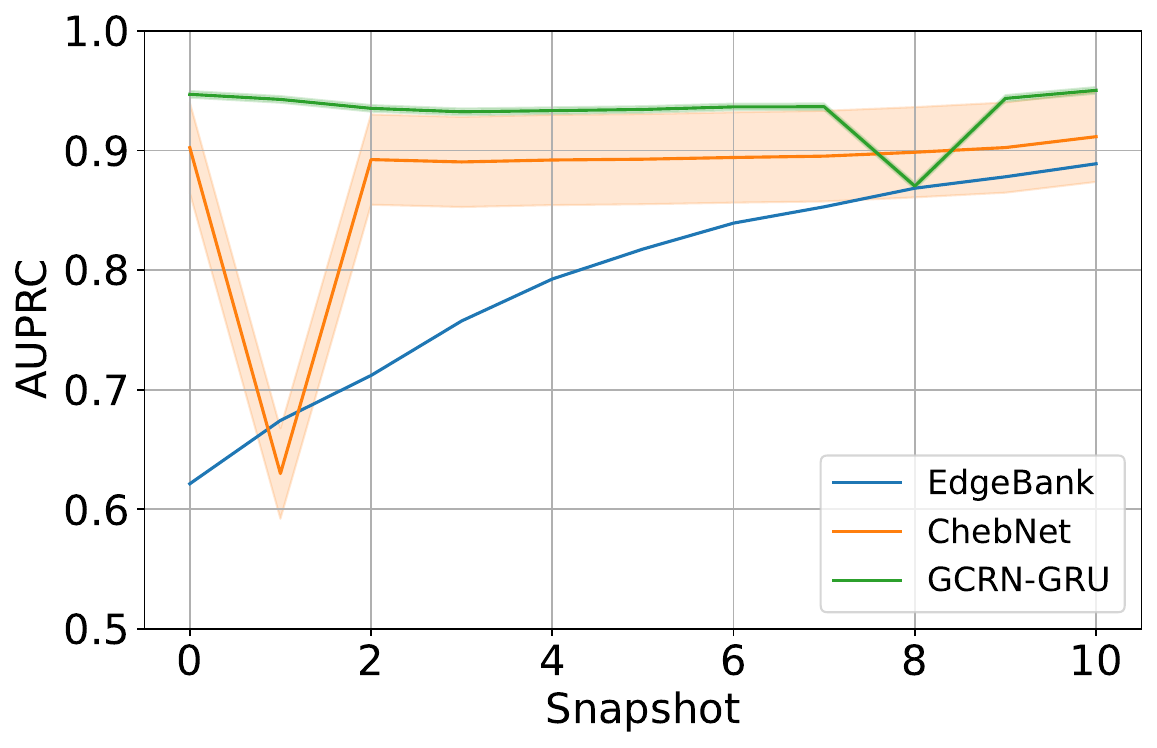}
\caption{}
\end{subfigure}
\caption{Performance trends over time for EdgeBank (blue line), Chebnet (orange line), and GCRN-GRU (green line) models for link prediction on mouse gastrulation (a), and pancreas (b).}
\label{fig:trend-lp}
\end{figure}

\paragraph{Link prediction.} We report the average AUPRC results in Table~\ref{tab:results_lp}. Overall, temporal graph learning approaches achieved the strongest performance across both datasets. In particular, GCRN-GRU consistently obtained the best average AUPRC on both mouse gastrulation and mouse pancreas, substantially outperforming standard message-passing architectures such as GCN and GAT. To better highlight the contribution of temporal modeling in GCRN-GRU compared to pure memorization and static graph approaches, we compare its temporal performance trend against EdgeBank and ChebNet in Figure~\ref{fig:trend-lp}. Results show that GCRN-GRU consistently maintains the best performance across developmental time, including later pseudotime snapshots where recurrency of edges is high. In contrast, the static ChebNet baseline exhibits larger fluctuations and lower performance over time.

\begin{table}[t]
\caption{Average AUPRC over time of the considered models for future link prediction on the gene regulatory networks. For each model, we ran experiments with 5 different random seeds, reporting the average result and standard deviation for each method.}
\label{tab:results_lp}
\centering
\begin{tabular}{lcc}
\toprule
Model & mouse-gastrulation & mouse-pancreas \\
\midrule
EdgeBank  & $0.88 \pm 0.00$ & $0.80 \pm 0.00$ \\
GCN       & $0.56 \pm 0.05$ & $0.56 \pm 0.04$ \\
GAT  & $0.62 \pm 0.07$ & $0.51 \pm 0.01$ \\
Chebnet  & $0.91 \pm 0.01$ & $0.88 \pm 0.09$ \\
EvolveGCN & $0.61 \pm 0.06$ & $0.64 \pm 0.04$\\
GCRN-GRU   & $\mathbf{0.95 \pm 0.01}$ & $\mathbf{0.94 \pm 0.03}$ \\
ROLAND (ChebConv+ GRU)    & $0.87 \pm 0.04$ & $0.86 \pm 0.02$\\
\bottomrule
\end{tabular}
\end{table}

\begin{figure}
    \centering
    \includegraphics[width=0.6\linewidth]{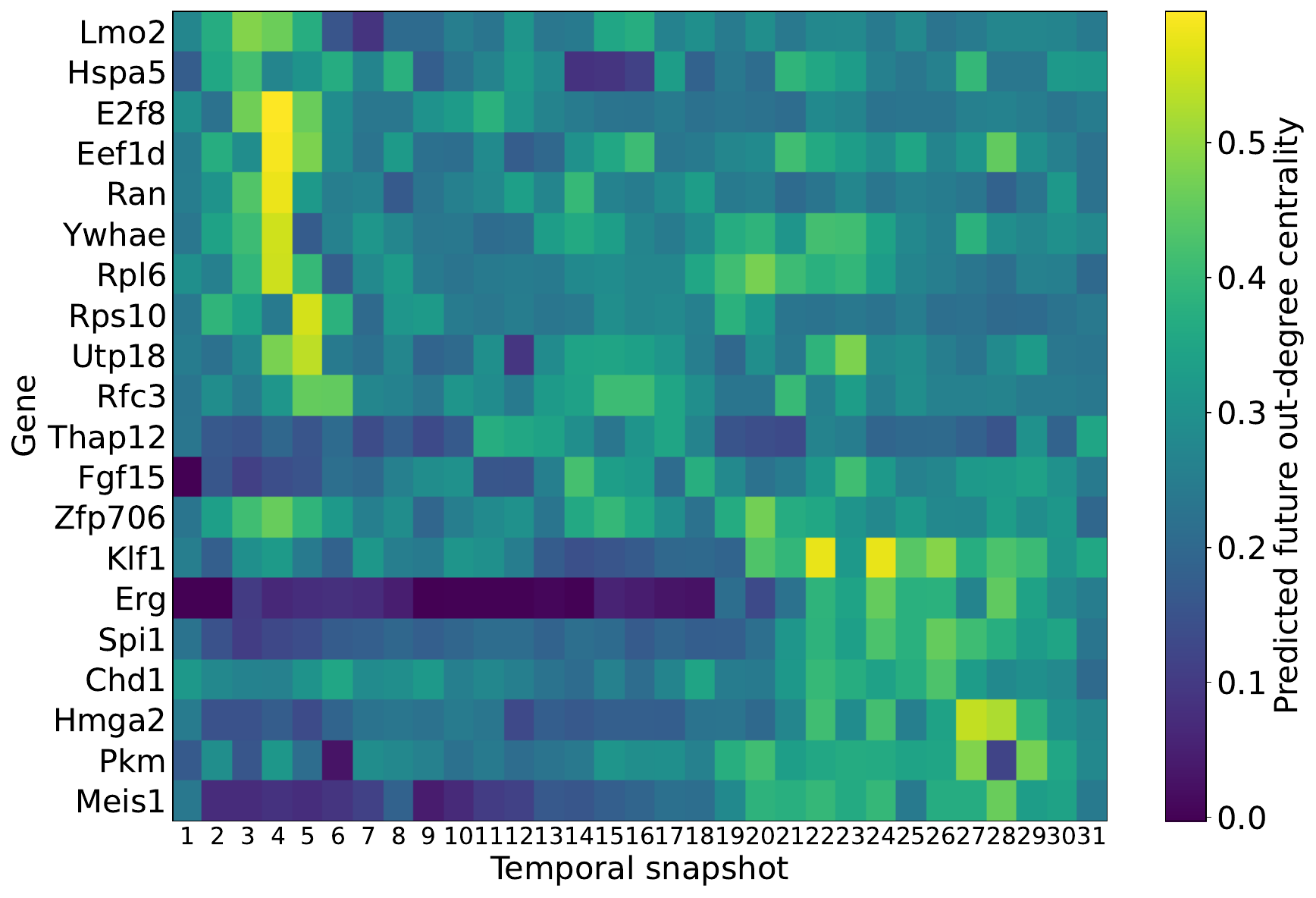}
    \caption{Temporal heatmap of the predicted future out-degree centrality for the top regulatory hubs identified by GCRN-GRU on mouse-gastrulation dataset.}
    \label{fig:hub-gastrulation}
\end{figure}

\paragraph{Forecasting future regulatory hubs.} We report the results in Table~\ref{tab:outdegree_forecasting}. Overall, spectral graph convolution architectures achieved substantially stronger performance than standard message-passing approaches across both datasets and metrics. To qualitatively investigate the biological relevance of the predicted regulatory hubs, we report in Figure \ref{fig:hub-gastrulation} the temporal heatmap of the predicted future out-degree centrality for the top regulatory hubs identified by the best-performing temporal graph learning model across developmental GRN snapshots of erythroid mouse gastrulation. To compute this heatmap, we first identify the top 20 highest predicted centralities over time, and then ask to forecast their complete temporal centrality profile to model checkpoints trained up to the previous temporal snapshot. The recovered trajectories were consistent with known hematopoietic and erythroid developmental programs. Early pseudotime stages were characterized by elevated predicted centrality for proliferative and biosynthetic regulators, including \textit{E2f8}, \textit{Ran}, and multiple ribosomal-associated genes. In contrast, later developmental stages showed increasing importance of canonical hematopoietic and erythroid regulators such as \textit{Klf1}, \textit{Spi1}, \textit{Erg}, and \textit{Meis1}, which are known to play key roles in erythroid differentiation and hematopoietic specification~\cite{Bieker2024klf1,scott1994pu1,taoudi2011erg,azcoitia2005meis1}. Overall, the temporal organization of the predicted hubs was coherent with known transitions underlying hematopoietic commitment and erythroid maturation.

\begin{table}[t]
\centering
\small
\caption{Comparison of methods for out-degree centrality forecasting on mouse gastrulation and mouse pancreas. For each model, we ran experiments with 5 different random seeds, reporting the average result and standard deviation for each method. Lower MAE is better, while higher Spearman and P@200 indicate better performance. Best results are in \textbf{bold}, second best are \underline{underlined}.}
\label{tab:outdegree_forecasting}
\setlength{\tabcolsep}{4pt}

\resizebox{\textwidth}{!}{
\begin{tabular}{lccc|ccc}
\toprule
& \multicolumn{3}{c|}{mouse-gastrulation}
& \multicolumn{3}{c}{mouse-pancreas} \\
\cmidrule(lr){2-4}
\cmidrule(lr){5-7}

Model
& MAE
& Spearman
& P@200
& MAE
& Spearman
& P@200 \\
\midrule

Linear
& 0.037 $\pm$ 0.00
& 0.009 $\pm$ 0.02
& 0.140 $\pm$ 0.02
& 0.034 $\pm$ 0.00
& -0.010 $\pm$ 0.02
& 0.110 $\pm$ 0.02 \\

MLP
& 0.040 $\pm$ 0.01
& 0.177 $\pm$ 0.08
& 0.229 $\pm$ 0.07
& 0.036 $\pm$ 0.00
& 0.106 $\pm$ 0.07
& 0.175 $\pm$ 0.05 \\

\midrule

GCN
& 0.037 $\pm$ 0.01
& 0.046 $\pm$ 0.06
& 0.156 $\pm$ 0.04
& 0.033 $\pm$ 0.00
& 0.055 $\pm$ 0.05
& 0.151 $\pm$ 0.03 \\

GAT
& 0.036 $\pm$ 0.01
& 0.084 $\pm$ 0.11
& 0.197 $\pm$ 0.11
& 0.033 $\pm$ 0.01
& 0.131 $\pm$ 0.08
& 0.207 $\pm$ 0.07 \\

ChebNet
& \underline{0.007 $\pm$ 0.01}
& \underline{0.496 $\pm$ 0.02}
& \underline{0.729 $\pm$ 0.06}
& \textbf{0.007 $\pm$ 0.01}
& \textbf{0.503 $\pm$ 0.01}
& \textbf{0.848 $\pm$ 0.05} \\

\midrule

EvolveGCN
& 0.036 $\pm$ 0.00
& 0.094 $\pm$ 0.08
& 0.190 $\pm$ 0.07
& 0.033 $\pm$ 0.00
& 0.082 $\pm$ 0.06
& 0.185 $\pm$ 0.05 \\

GCRN-GRU
& \textbf{0.005 $\pm$ 0.00}
& \textbf{0.499 $\pm$ 0.02}
& \textbf{0.741 $\pm$ 0.05}
& \textbf{0.007 $\pm$ 0.01}
& 0.492 $\pm$ 0.02
& 0.808 $\pm$ 0.05 \\

ROLAND (ChebConv+ GRU)
& 0.020 $\pm$ 0.01
& 0.485 $\pm$ 0.04
& 0.713 $\pm$ 0.07
& 0.012 $\pm$ 0.01
& \underline{0.501 $\pm$ 0.01}
& \underline{0.832 $\pm$ 0.03} \\

\bottomrule
\end{tabular}
}
\end{table}

\section{Conclusion}
In this work-in-progress paper, we investigate temporal graph learning as a framework for modeling and predicting the evolution of biological systems from single-cell transcriptomic data. We represent developmental cellular states as pseudotime-resolved gene regulatory networks and model them as evolving graph structures over persistent gene identities. This formulation enabled the application of temporal graph neural networks to developmental single-cell systems despite the absence of explicit temporal observations of individual cells. We evaluate this framework on mouse erythroid gastrulation and pancreas endocrinogenesis considering three complementary forecasting tasks: future gene-expression prediction, future link prediction, and future out-degree centrality prediction. Across tasks, temporal graph learning approaches consistently achieved strong performance, with spectral graph architectures such as GCRN-GRU, emerging as particularly effective. For gene-expression forecasting, graph-based models outperformed biological foundation-model baselines such as scGPT and scFoundation, suggesting that explicitly modeling evolving regulatory structure provides useful information beyond static pretrained representations. For link prediction and future hub forecasting, temporal graph learning captured non-trivial regulatory dynamics and enabled the identification of temporally important regulatory genes. Overall, our findings support temporal graph learning as a promising and complementary paradigm to current foundation-model approaches for modeling dynamic biological systems. 
Future work will extend this benchmark to additional developmental datasets, biological foundation models, and more recent temporal attention-based architectures for dynamic graph learning.

%
%
\bibliographystyle{unsrt}
\bibliography{bibliography}

\section*{Appendix}

\subsection*{Additional information on Dataset}

\begin{figure*}[t]
\centering

\begin{subfigure}[t]{0.48\textwidth}
\centering
\includegraphics[width=0.95\linewidth,height=4cm]{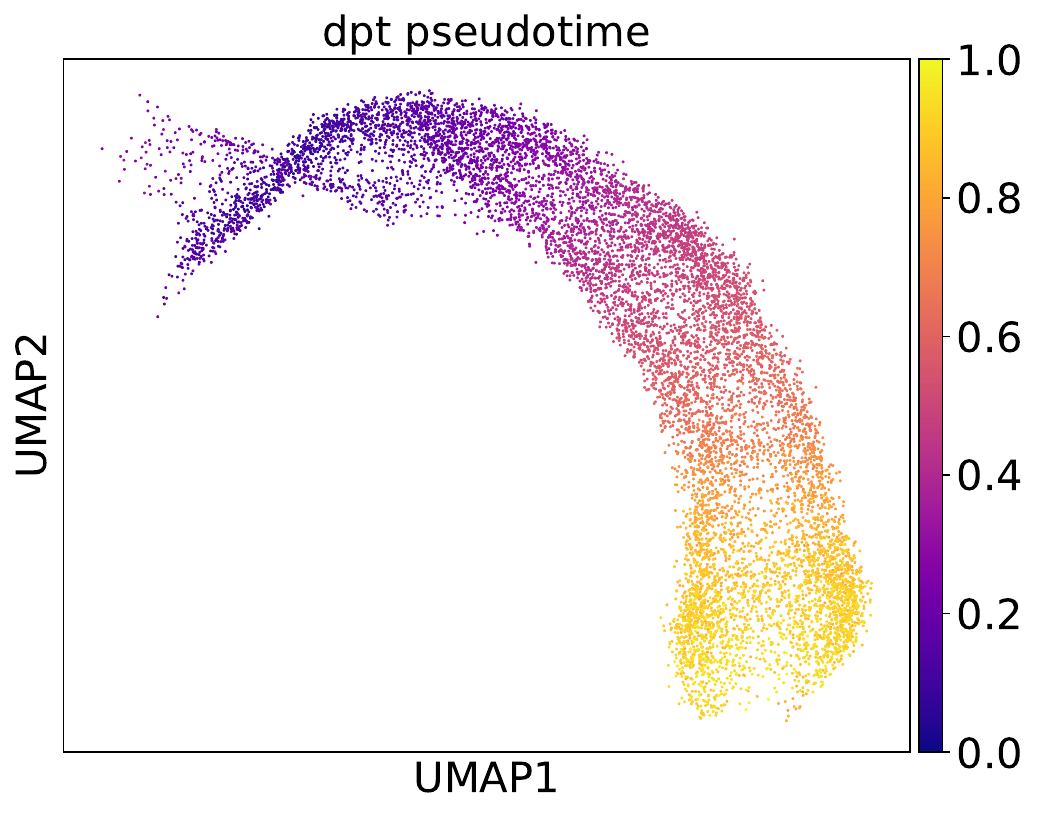}
\caption{}
\end{subfigure}
\hfill
\begin{subfigure}[t]{0.48\textwidth}
\centering
\includegraphics[width=0.95\linewidth,height=4cm]{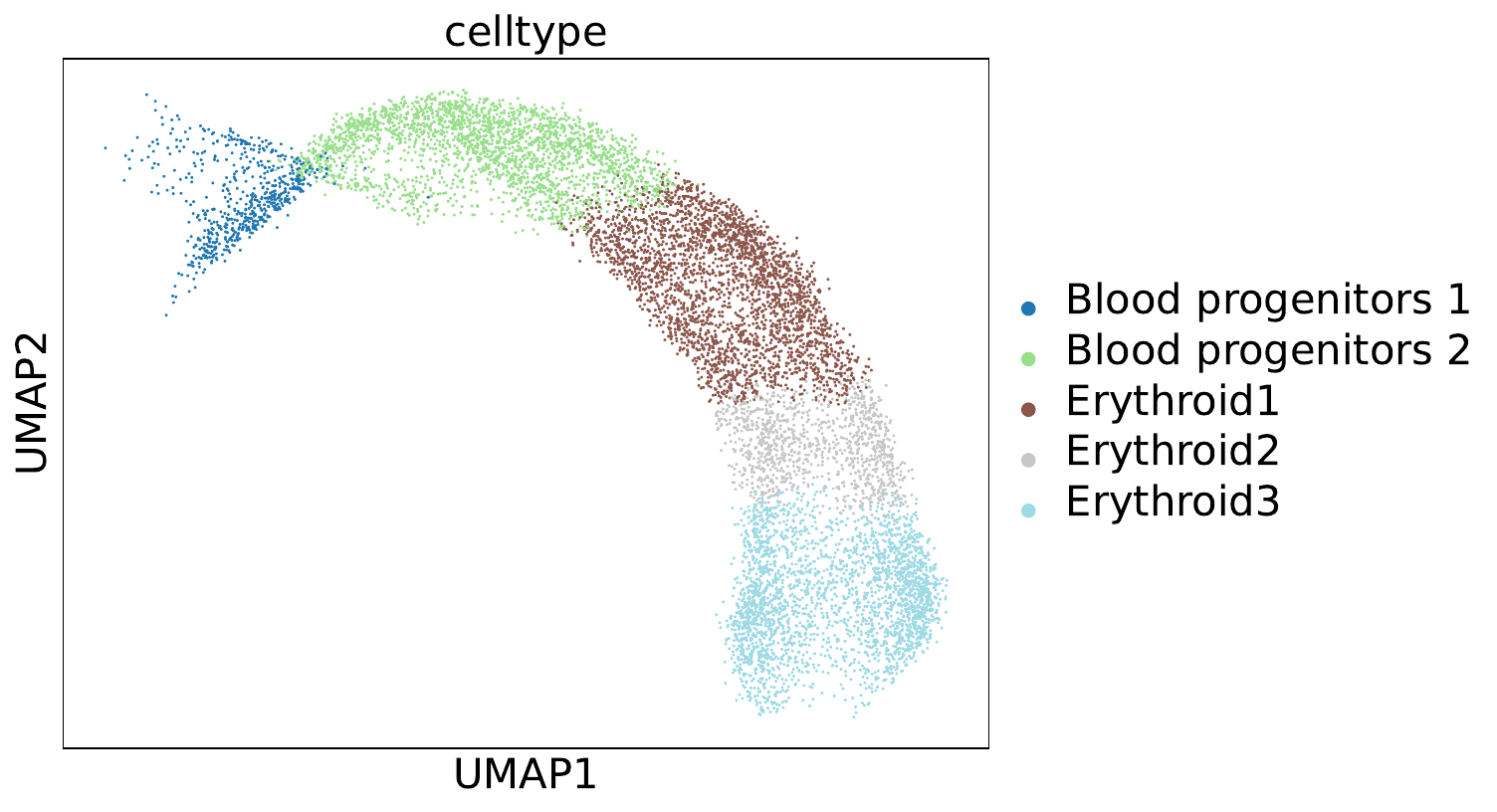}
\caption{}
\end{subfigure}

\vspace{0.35cm}

\begin{subfigure}[t]{0.48\textwidth}
\centering
\includegraphics[width=0.95\linewidth,height=4cm]{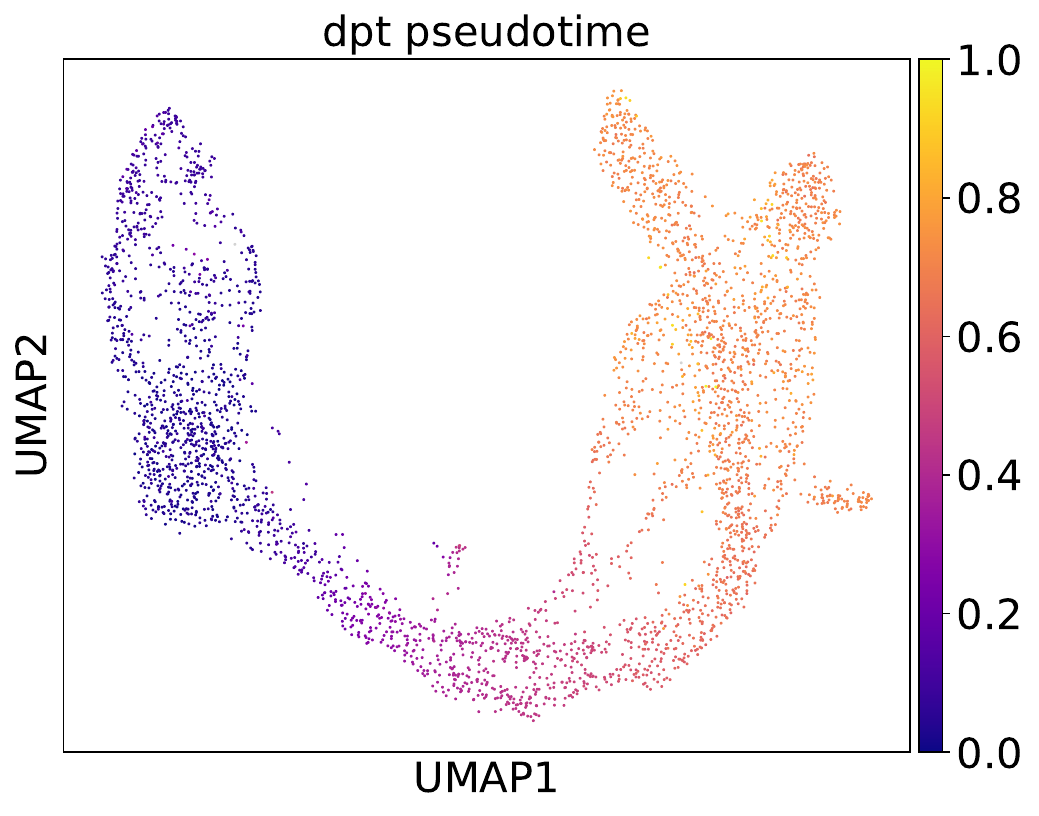}
\caption{}
\end{subfigure}
\hfill
\begin{subfigure}[t]{0.48\textwidth}
\centering
\includegraphics[width=0.95\linewidth,height=4cm]{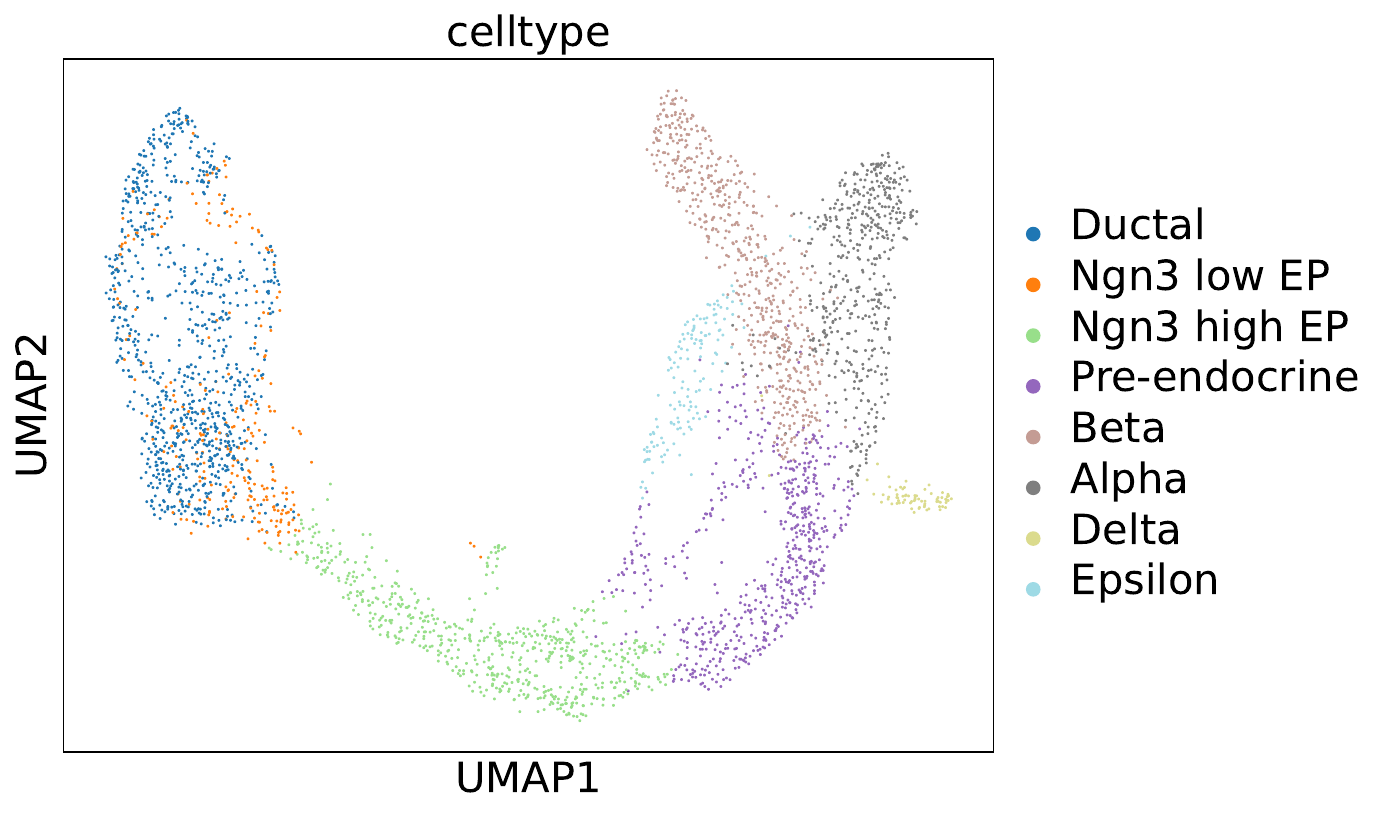}
\caption{}
\end{subfigure}

\caption{Diffusion pseudotime and cell-type UMAP visualization for mouse gastrulation (a)--(b) and pancreas (c)--(d) datasets.}
\label{fig:pseudotime}
\end{figure*}

\paragraph{Pseudotime organization of developmental trajectories.}
We report in Figure~\ref{fig:pseudotime} the UMAP visualization of the gastrulation and pancreas single-cell datasets together with the corresponding diffusion pseudotime estimates used to construct temporal GRN snapshots.
In both datasets, the inferred pseudotime coordinates were consistent with the underlying cellular organization and revealed smooth developmental progressions across transcriptional states. The gastrulation dataset displayed a continuous transition from blood progenitors toward erythroid populations, whereas the pancreas dataset showed progressive organization across endocrine and epithelial cell populations. These results indicate that diffusion pseudotime provides a biologically coherent temporal ordering of cells and supports the construction of pseudotime-resolved gene regulatory networks for downstream temporal graph learning analyses.

\section{Additional information on models}

We considered a broad set of graph-based, temporal, tabular, and biological foundation-model baselines. In the following, we briefly summarize the considered architectures.

\paragraph{Static graph neural networks.}
We first considered standard graph neural network architectures.

\begin{itemize}
    \item \textbf{GCN} \cite{gcn}: Graph Convolutional Networks perform message passing through neighborhood aggregation based on a normalized graph Laplacian operator. Each layer updates node representations by combining information from neighboring nodes and applying a learned linear transformation.

    \item \textbf{GAT} \cite{gat}: Graph Attention Networks extend standard graph convolutions through attention mechanisms that assign learnable importance weights to neighboring nodes during message aggregation. This allows the model to adaptively focus on the most relevant local interactions.

    \item \textbf{ChebNet} \cite{chebnet}: Chebyshev Networks perform spectral graph convolutions through truncated Chebyshev polynomial approximations of the graph Laplacian. Unlike standard message-passing approaches, ChebNet naturally incorporates information from multi-hop neighborhoods within a single layer while avoiding explicit recursive propagation over immediate neighbors only.
\end{itemize}

\paragraph{Temporal graph neural networks.}
We adopt three state-of-the-art discrete-time temporal graph neural networks as primary temporal baselines.

\begin{itemize}
    \item \textbf{EvolveGCN} \cite{evolvegcn}: EvolveGCN captures the dynamics of graph sequences by evolving the parameters of graph convolutional layers through recurrent neural networks. Instead of directly learning static GNN parameters, the model uses recurrent updates to adapt the graph convolution weights over time, enabling temporal parameter evolution across snapshots.

    \item \textbf{GCRN-GRU} \cite{gconvgru}: GCRN-GRU generalizes recurrent neural networks to graph-structured data by replacing the linear transformations inside GRU cells with graph convolution operators. In our implementation, spatial information is modeled through Chebyshev spectral convolutions~\cite{chebnet}, while recurrent gating mechanisms capture temporal dependencies between consecutive GRN snapshots.

    \item \textbf{ROLAND} \cite{roland}: ROLAND is a framework for extending static GNNs to dynamic graph settings through explicit node-state evolution across time. Node embeddings at different graph convolutional layers are interpreted as hierarchical node states, which are updated as new graph snapshots are observed. In this work, message passing is implemented using Chebyshev spectral convolutions~\cite{chebnet}, while temporal updates are modeled through gated recurrent units (GRUs), which has been shown as the most effective node update solution for discrete-time temporal networks \cite{roland}.
\end{itemize}

\paragraph{Biological foundation models.}
We additionally considered recent large-scale biological foundation models pre-trained on massive collections of single-cell transcriptomic data.

\begin{itemize}
    \item \textbf{scGPT} \cite{scgpt}: scGPT is a generative transformer model for single-cell biology inspired by large language models. The model learns contextualized representations through autoregressive and masked gene-expression prediction objectives, primarily producing cell-level embeddings that capture global transcriptional states across cells. To adapt scGPT to temporal GRN forecasting tasks, we first compute pretrained cell embeddings for all cells and subsequently aggregate them /by average) within each pseudotime snapshot to obtain a snapshot-level embedding representation. This pooled embedding is then concatenated to node-level gene features and used as additional contextual information for downstream temporal graph learning tasks.

    \item \textbf{scFoundation} \cite{scfoundation}: scFoundation is a large-scale pretrained transformer architecture specifically designed for large-scale single-cell representation learning. Unlike scGPT, which primarily focuses on contextual gene prediction, scFoundation emphasizes scalable representation learning across tens of millions of cells through hierarchical embedding strategies and large-scale masked reconstruction objectives. It additionally provides gene-level representations, making it particularly suitable for gene-level downstream tasks. To adapt scFoundation to temporal GRN forecasting, we compute pretrained gene embeddings across cells and aggregate them within each pseudotime snapshot to obtain snapshot-specific gene representations, which are subsequently used as node-level features for downstream forecasting tasks.
\end{itemize}

For experiments on mouse datasets, both scGPT and scFoundation were applied after converting mouse genes to their corresponding human orthologs using MousePy \footnote{\url{https://newcraftgroup.github.io/mousepy/index.html}, June 2026.}, allowing compatibility with the human-pretrained vocabularies of the considered biological foundation models while preserving alignment with the original GRN node identities.

\end{document}